\documentclass[a4paper,twoside]{article}

\usepackage{epsfig}
\usepackage{subcaption}
\usepackage{calc}
\usepackage{amssymb}
\usepackage{amstext}
\usepackage{amsmath}
\usepackage{amsthm}
\usepackage{multicol}
\usepackage{pslatex}
\usepackage[bottom]{footmisc}

\usepackage{natbib}
\usepackage{todonotes}
\usepackage{setspace}
\usepackage{hyperref}

\usepackage{rotating}
\usepackage[export]{adjustbox}

\usepackage{SCITEPRESS}     

\begin{document}

\title{PanDepth: Joint Panoptic Segmentation and Depth Completion}

\author{\authorname{Juan Pablo Lagos\sup{1}, Esa Rahtu\sup{1}\orcidAuthor{0000-0001-8767-0864}, }
\affiliation{\sup{1}Tampere University, Tampere, Finland}
\email{\{juanpablo.lagosbenitez, esa.rahtu \}@tuni.fi}
}



\abstract{Understanding 3D environments semantically is pivotal in autonomous driving applications where multiple computer vision tasks are involved. Multi-task models provide different types of outputs for a given scene, yielding a more holistic representation while keeping the computational cost low. We propose a multi-task model for panoptic segmentation and depth completion using RGB images and sparse depth maps. Our model successfully predicts fully dense depth maps and performs semantic segmentation, instance segmentation, and panoptic segmentation for every input frame. Extensive experiments were done on the Virtual KITTI 2 dataset and we demonstrate that our model solves multiple tasks, without a significant increase in computational cost, while keeping high accuracy performance. Code is available at \url{https://github.com/juanb09111/PanDepth.git}.
}

\onecolumn \maketitle \normalsize \setcounter{footnote}{0} \vfill

\section{\uppercase{Introduction}}
\label{sec:introduction}
Producing a holistic representation of a given scene has become essential in computer vision. The traditional tasks and challenges, such as semantic segmentation, instance segmentation, pose estimation, edge estimation, or depth completion only provide a limited representation that alone are not enough to successfully complete more complex tasks, for instance, autonomous driving, where, in addition to estimating the distance of the objects and stuff on and around the road, it is also essential to understand the semantic context of the scene, that is, identifying the type of objects around, e.g. cars, pedestrians, road lanes, traffic signs, at the same time as the depth to such objects is estimated. This raises the need for multi-task models that are capable of solving several tasks in parallel while keeping the computational cost low.

\begin{figure}[ht]
    \centering
    \includegraphics[width=0.45\textwidth]{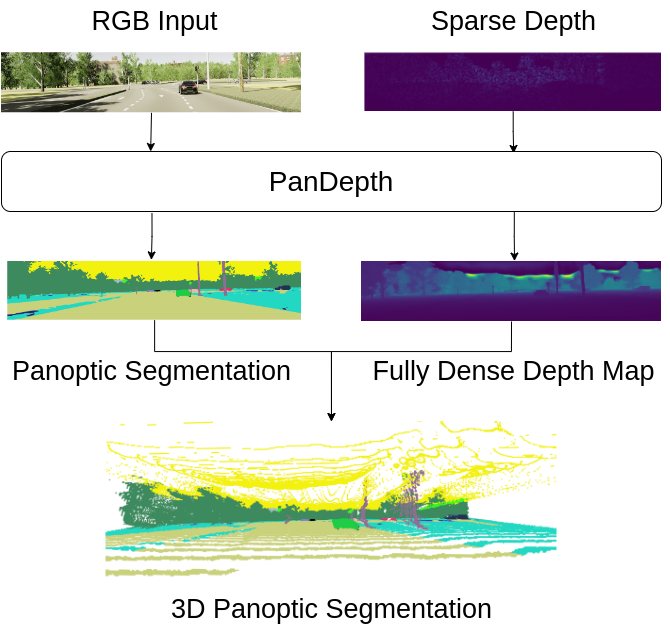}
    \caption{The proposed model (PanDepth) takes RGB images and sparse depth and returns the corresponding panoptic segmentation and fully dense depth map with which we create a 3D panoptic segmentation representation of the input frame.}
    \label{fig:overview_0}
\end{figure}
This work is inspired by the idea  of devising a model that combines panoptic segmentation and depth completion which is of high relevance in applications such as autonomous driving where understanding 3D environments semantically is pivotal for the performance of autonomous machines. We explore the hypothesis that panoptic segmentation and depth completion can use cues from one another, more explicitly, that there are depth features that contain relevant semantic cues as well there are semantic segmentation features that contain relevant depth cues.

Multi-task networks, not only reduce the demand for computational resources, as compared to running multiple single-task networks but also, there is empirical evidence that multi-task networks can perform better in each individual task by jointly learning features from all tasks involved \citep*{mtask, mtask2, lagos}. For instance, depth features can be helpful for performing semantic segmentation and vice-versa. Even in applications where solving a single task is the primary goal, introducing features from other tasks may leverage accuracy and performance. Such paradigm is known as auxiliary tasks \citep*{liebel2018auxiliary, aux2}, whereby solving other tasks it is possible to obtain relevant features which lead to a better performance in the main task.

We focus on solving three tasks in a joint manner, namely, semantic segmentation, instance segmentation, and depth completion using convolutional neural networks (CNNs). Combining semantic segmentation and instance segmentation into one single representation is known as panoptic segmentation \citep{panoptic}. It provides a representation of an image where not only every pixel is assigned a label from a list of predefined labels, as in the case of semantic segmentation, but also, objects are detected as instances of a specific class, thus providing valuable information, such as the number of cars, people, or objects of a certain kind that are found in the image, as well as the semantic context of the non-countable stuff in the scene. Countable and non-countable objects are usually referred to as "things" and "stuff" in the context of computer vision \citep{Adelson2001OnSS}. 

While semantic segmentation produces a single output, pixel-wise classification, instance segmentation produces three different outputs: bounding boxes for the objects detected, a label for each bounding box, and a segmentation mask for each object detected. The outputs of both tasks, semantic segmentation, and instance segmentation, are usually fused using heuristic methods with no learnable parameters \citep*{effps, UPS}. 

On the other hand, depth completion aims to produce a dense depth map from sparse depth points which cover only a few pixels from a given image. Sparse depth maps  can be obtained with active depth sensors, such as lidars. When 3D points obtained with lidars are projected onto an image, only about $5\%$ of the image is covered \citep{sparsity}. The goal is then to produce a dense depth map, with depth values for all the pixels in the image, given a sparse depth map as input.

In this paper, we propose an end-to-end model for panoptic segmentation and depth completion using joint training in order to provide a more holistic representation of the input images. In contrast with other works where predictions are made based on RGB images only \citep*{PanopticDepth, Schon_2021, PolyphonicFormer}, our model processes heterogeneous data jointly, that is, RGB images and sparse depth maps as shown in Figure \ref{fig:overview_0}. For most machine perception applications, active depth sensors are part of the setup, for which we consider it more relevant to integrate both RGB images as well as sparse depth maps. We also quantify the effects of joint training as compared to training every task individually, thus providing more data on the growing evidence of the advantages of multi-task networks. We conduct extensive experiments on  Virtual KITTI 2 \citep{cabon2020virtual}, which is a relevant dataset in the context of autonomous driving that contains ground truth annotations for instance segmentation, semantic segmentation and ground truth depth maps available for the entire dataset. Although panoptic segmentation ground truth is not directly provided by Virtual KITTI 2 dataset, we use semantic and instance segmentation ground truth to generate panoptic segmentation annotations.

\section{\uppercase{Related Works}}

\subsection{Panoptic Segmentation}

Early works in computer vision developed CNN architectures for performing semantic segmentation and instance segmentation independently with reasonable success \citep*{DBLP, unet, maskrcnn}. Later on, \citet*{panoptic} proposed a task that would combine both tasks into one, which they named panoptic segmentation. \citet*{panoptic} also defined a metric for assessing the performance of panoptic segmentation predictions referred to as panoptic quality (PQ), thus, providing a complete definition of the problem of panoptic segmentation with a target metric for performance comparison. Such a robust definition of the task called the attention of the community, leading to the first architectures for end-to-end panoptic segmentation using CNNs \citep*{attention_guided, rt_pan, panoptic_deeplab, UPS, end2, fast, panoptic_fpn, autod}. 

The most common challenges that appeared with panoptic segmentation are how to optimize a shared feature extractor as well as how to combine semantic segmentation and instance segmentation predictions while keeping the computational cost low. \citet{effps} proposed a model for panoptic segmentation which consists of two heads, namely, semantic segmentation and instance segmentation, a fusion module for combining the outputs of both heads, and a feature extractor based on a family of scalable CNNs known as EfficientNet \citep{effnet}, where the resolution, depth, and width are balanced depending on the computational resources available. For multi-scale features, \citet{effps} wrap the feature extractor into a two-way feature pyramid network (FPN). Similarly, \citet{sc} used the same concept of scalable networks to Residual Networks (ResNets) for performing panoptic segmentation.

Other approaches \citep*{pant1, detr, detr2, maskt, pant2, mask2former} have adopted transformers architecture \citep{attention}, initially designed for text processing and sequence transduction, and integrated attention mechanisms for panoptic segmentation. In contrast with more traditional methods, with instance segmentation and semantic segmentation defined as sub-tasks, transformer-based models use queries to represent "things" and "stuff" classes and perform panoptic segmentation.

\subsection{Depth Completion}

The task of depth completion aims to transform a sparse depth map, usually obtained with active depth sensors e.g. Lidar, into a dense depth map. Lidar devices can only provide a limited amount of depth points when projected onto the corresponding image, raising the need for methods that can lead to a fully-dense representation of the depth of an entire image. Several works have used RGB images as guidance for depth completion \citep*{deeplidar, confprop, spnoisy, guidedconv, DDP, nonloc, PEnet}. \citet{depthandsem} proposed an encoder-decoder network architecture for depth completion, based on a late fusion of RGB images and sparse depth maps. However, processing RGB and Lidar data is not trivial, since, in contrast to RGB images, sparse depth data lacks a natural grid structure unless projected onto a 2D space which also facilitates the usage of traditional 2D convolutional layers. 

Nonetheless, when mapping 3D data to 2D, valuable information regarding the geometrical relationship among the points in the 3D space is lost. \citet{2D3D} introduced a fuse block that exploits 3D cues by using parametric continuous convolution layers \citep{Wang_2018} while using 2D convolutions for processing RGB and later fusing the corresponding features in 2D space. Such 2D-3D fuse method is an essential building block in the proposed model, in which, with slight modifications to the model proposed by \citet{2D3D}, we successfully  map sparse depth maps to dense depth maps.

\subsection{Multi Task Learning}
CNNs can benefit from performing multiple tasks, as opposed to single-task networks. Branched CNNs consist of shared layers as well as task-specific layers, also known as branches. When such CNNs are trained, the weights of the shared layers are adjusted via back-propagation from each one of the branches, each one of which has one or multiple loss functions defined. In turn, the shared layers learn relevant features for all tasks, and such features are then fed to every branch. That allows for a very distinctive flow of information between the different branches. There is increasing evidence that single tasks, benefit when models are trained jointly improving the performance of each one of the tasks tackled by the network \citep*{auxiliary, multiatt, multidepth, simsemseg, ltb}.

While some multi-task networks have addressed tasks relatively similar e.g. instance segmentation and semantic segmentation, other works have combined semantic segmentation and depth completion as end-to-end models \citep*{hazirbas16fusenet, Simultaneous, SOSD}. \citet{lagos} proposed a combined model for semantic segmentation and depth completion using RGB images and sparse depth maps, where it is demonstrated quantitatively and visually how each task outperforms equivalent single-task models for semantic segmentation and depth completion trained independently. Our model performs depth completion, instance segmentation, and semantic segmentation. We fuse instance and semantic segmentation to obtain a panoptic segmentation representation. In contrast with other methods, our model processes heterogeneous data, more specifically RGB images, and sparse depth maps using a stack of 2D-3D fuse blocks as proposed by \citet{2D3D}

\section{\uppercase{Architecture}}

\subsection{Overview}

The proposed model performs panoptic segmentation and depth completion in an end-to-end manner. It consists of a two-way feature pyramid network (FPN) as a shared feature extractor, three task-specific branches, one for each task (semantic segmentation, instance segmentation, and depth completion), one joint branch that refines the semantic logits using the resulting depth maps as guidance, and one final block for combining semantic and instance logits based on the fusion block proposed by \citet{effps}. The inputs to our model are RGB images and sparse depth maps and the output is the corresponding panoptic segmentation representations and fully dense depth maps. The panoptic segmentation output and the resulting depth maps can be further combined to produce a 3D panoptic segmentation representation as shown in Figure \ref{fig:overview_0}.

\begin{figure}
    \centering
    \includegraphics[width=0.45\textwidth]{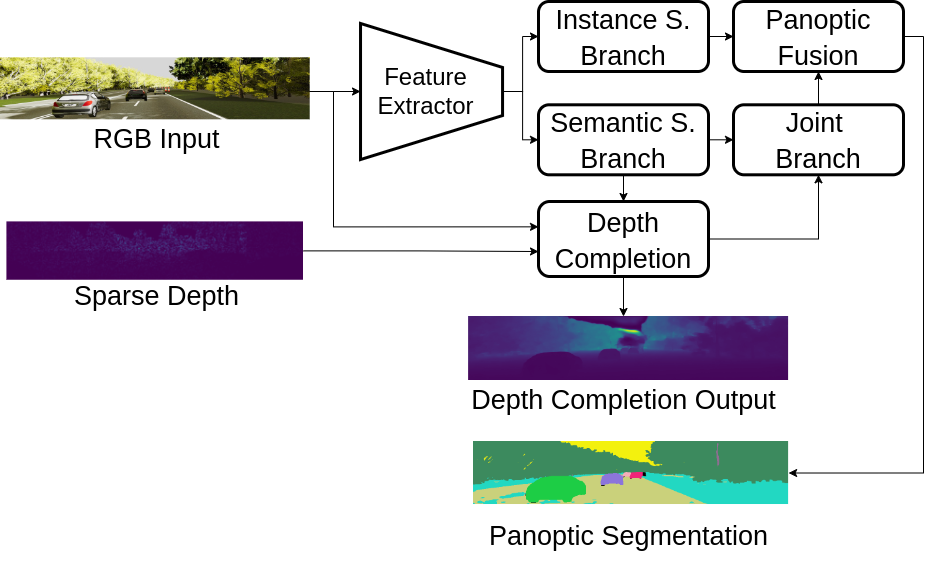}
    \caption{Overview of the proposed PanDepth architecture. Given an RGB image and sparse depth map as input, our model outputs the corresponding dense depth map and panoptic segmentation.}
    \label{fig:overview}
\end{figure}

\begin{figure*}[ht]
        \centering
        \includegraphics[width=1\textwidth]{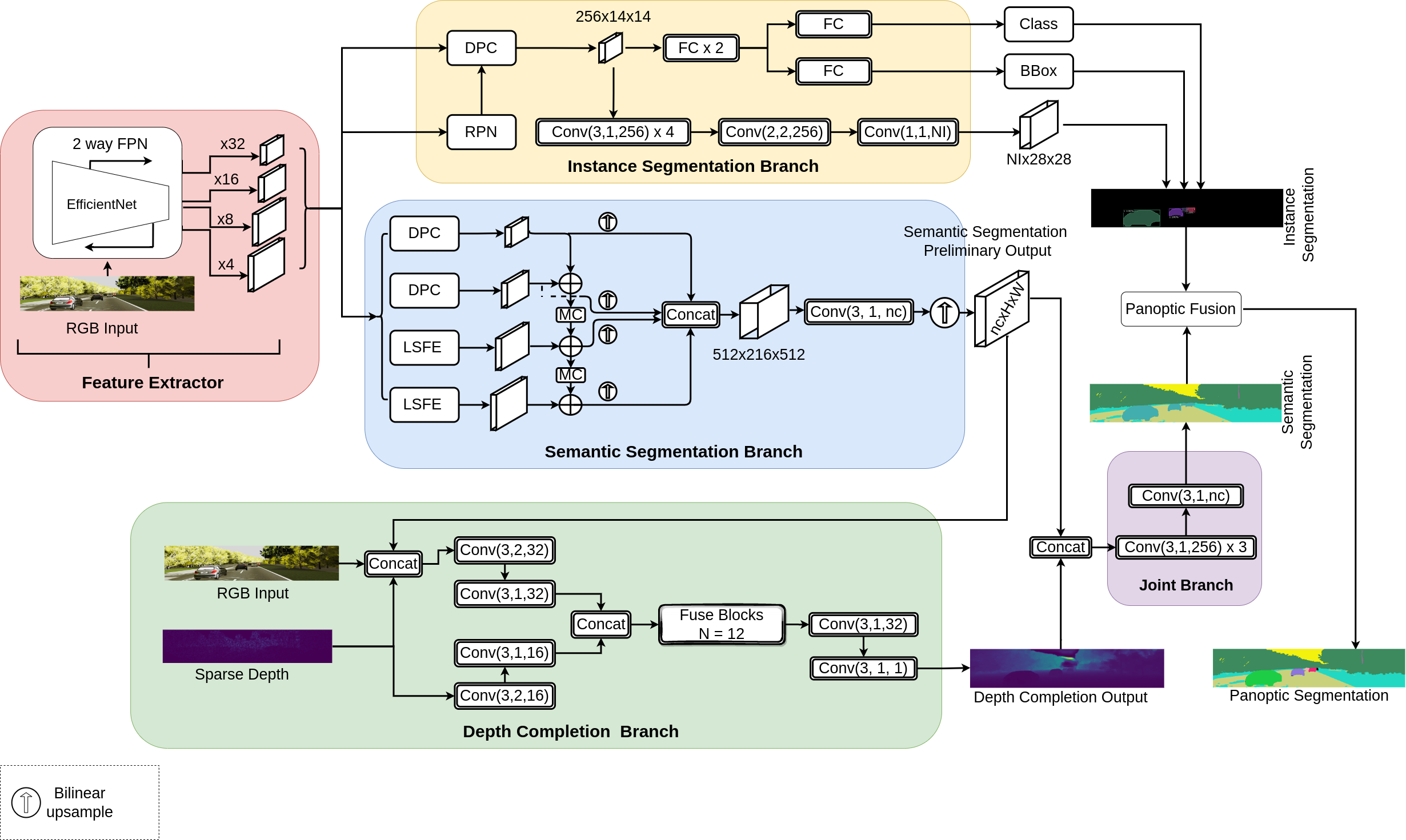}
        \caption{PanDepth architecture. Our model consists of a feature extractor, three task-specific branches (\textit{i.e.} instance segmentation, semantic segmentation, and depth completion), a joint branch, and a panoptic fusion module. The convolutional layers in this diagram follow the notation Conv($k$,$s$,$c$) $\times n$ representing a stack of $n$ convolutional layers where $k$ refers to a kernel of size $k \times k$, $s$ is the stride, $c$ is the number of output feature channels, and $FC$ represents a fully connected layer.}
    
    \label{fig:pan_depth}
\end{figure*}

\subsection{Backbone}

The backbone consists of a two-way FPN with an EfficientNet-B5 \citep{effnet} at the core, as shown in Figure \ref{fig:pan_depth}. On one hand, the FPN upsamples lower resolution features and adds them together. On the other hand, the FPN downsamples higher-resolution features and adds them together. This allows for multi-scale feature extraction. The backbone returns feature maps at four different scales, downscaled by a factor of $\times 4$, $\times 8$, $\times 16$, and $\times 32$ with respect to the spatial resolution at the input.

\subsection{Semantic Segmentation Branch}

The semantic segmentation branch is a light-weighted structure that consists of three main building blocks based on the model proposed by \citet{effps}. Firstly, a Large Scale Feature Extractor (LSFE)  extracts localized fine features. Secondly, a small-scale feature extractor based on Dense Predictions Cells (DPC), and finally, a Mismatch Correction Module (MC) is used in order to properly aggregate features at different scales. The input to this branch consists of the four feature maps returned by the backbone, they are in four different scales, $\times 4$, $\times 8$, $\times 16$, and $\times 32$. The tensors returned by the LSFE and DPC modules are aggregated as shown in Figure \ref{fig:pan_depth}. Finally, this branch returns preliminary semantic segmentation logits of size $nc \times H \times W$, where $nc$ is the total number of classes and $H \times W$ refers to the spatial resolution of the input $height \times width$ respectively. At a later stage, the preliminary semantic segmentation logits are refined with depth maps as guidance in the joint branch.

\subsection{Instance Segmentation Branch}

The instance segmentation branch is a lighter version of Mask R-CNN \citep{maskrcnn}. Following the modifications suggested by \citet{effps}, all the convolutions were replaced by depthwise separable convolutions \citep{xception}, batch normalization layers were replaced by synchronized Inplace Activated Batch Normalization layers (iABN) \citep{iabn} and the ReLU activations were replaced by Leaky ReLU.

Similar to Mask R-CNN, the instance segmentation branch consists of two stages. In the first stage, a region proposal network (RPN) returns a set of rectangular regions with a corresponding objectness score. Thereafter, a RoIAlign module extracts small feature maps of size $7 \times 7$ from the regions returned by the RPN. Subsequently, those features are used as input to two sub-branches that run in parallel, one of which regresses bounding boxes and classifies the objects of each corresponding box, and another sub-branch that regresses the corresponding masks returning an output tensor of size $NI \times 28 \times 28$, where $NI$ corresponds to the number of instances detected.

\subsection{Depth Completion Branch}\label{sec:dc}

This branch processes frame by frame and takes three different input types. Firstly, a sparse depth map originated from a $3D$ to $2D$ projection of a point cloud. Secondly, the corresponding RGB frame, and thirdly a preliminary semantic segmentation map as shown in Figure \ref{fig:pan_depth}. Our depth completion branch is based on the architecture proposed by \citet{chen2020learning}, upon which we made modifications in order to use preliminary semantic segmentation maps as proposed by \citet{lagos}. At the input level, the sparse depth map is passed through two 2D convolutional layers of kernel size $3 \times 3$, while the RGB image and the semantic segmentation map are concatenated and passed through two $2D$ convolutional layers of kernel size $3 \times 3$. Subsequently, the two corresponding outputs are concatenated and, along with the original sparse depth map, they serve as input to a stack of $N$ $2D-3D$ Fuse Blocks. Finally, the resulting tensor from the Fuse Blocks passes through two $2D$ convolutional layers of kernel size $3 \times 3$ for refinement, yielding the final fully dense depth map.

\subsection{Joint Branch}

We use the depth completion output as guidance for refining the semantic segmentation preliminary output. This branch, albeit simple, successfully leverages the performance of the semantic segmentation task. It consists of four stacked $2D$ convolutions of kernel size $3 \times 3$. The input to this branch is the concatenation of the output of the depth completion branch, that is, a fully dense depth map, and the preliminary semantic segmentation output. Finally, this branch returns a tensor of size $nc \times H \times W$, where $nc$ is the total number of classes in the dataset, $H$ and $W$ correspond to the original height and width of the model's input respectively.

\subsection{Loss Functions}

\paragraph{Semantic Segmentation}\label{semseg_sec}

We used the weighted per-pixel log-loss for semantic segmentation. It is defined as follows:

\begin{equation}\label{eq1}
    L_{semantic} = -\sum_{i}w_i p_i \log \hat{p_i},
\end{equation}

where $i$ is the pixel index, $w_i = \frac{4}{WH}$ if pixel $i$ is within the $25 \%$ worst predictions, $w_i = 0$ otherwise. $W$ and $H$ correspond to the width and height of the input image respectively, $p_i$ and $\hat{p_i}$ are the ground truth and the predicted probability for pixel $i$ of belonging to class label $c \in p$ respectively. The predicted probability $\hat{p_i}$ is computed using the Softmax function defined as:

\begin{equation}\label{eq2}
    Softmax(x_n) = \frac{exp(x_n)}{\sum_{m} exp(x_m)}.
\end{equation}

\paragraph{Instance Segmentation}\label{instanceseg_sec}

We adopted the loss functions for instance segmentation as defined in Mask R-CNN \citep{maskrcnn}. There are loss functions defined for the two stages of this branch. In the first stage (the RPN), we calculate two losses, namely, objectness score loss ($L_{os}$) and object proposal loss ($L_{op}$. For the second stage, we calculate three losses, classification loss $L_{cls}$, bounding-box regression loss $L_{box}$, and mask loss $L_{mask}$. The total loss for the instance segmentation branch is given by:

\begin{equation}\label{eq_instance}
    L_{instance} = L_{os} + L_{op} + L_{cls} + L_{box} + L_{mask}
\end{equation}

\paragraph{Depth Completion}

We used Mean Squared Error (MSE) as loss for the depth completion branch. The MSE was calculated and averaged over the pixels for which the corresponding ground truth depth values were available in the sparse depth map. The loss function is defined by

\begin{equation}\label{eq3}
    L_{depth} = \frac{1}{N}\sum_{i}(\hat{y_i} - y_i)^2,
\end{equation}

where $N$ is the number of pixels, $\hat{y_i}$ is the predicted value and $y_i$ is the ground truth value for pixel $i$.

\paragraph{Joint Loss} In addition to the loss function related to each task, we compute a loss involving each one of the tasks performed by our model, in particular, semantic segmentation, instance segmentation, and depth completion. This loss is simply the sum of every specific loss as in e.q. \ref{eq_joint}

\begin{equation}\label{eq_joint}
    L_{joint} =  L_{semantic} + L_{instance} + L_{depth}.
\end{equation}

\section{Experiments}

\subsection{Implementation Details}

We trained our model for 50 epochs on one machine with four 32GB  graphics processing units (GPUs) running in parallel.  The loss functions were optimized using Adam algorithm with a learning rate set to $0.0002$.

\subsection{Dataset} 

 We trained and tested our models on Virtual KITTI 2  \citep{cabon2020virtual}. It is a synthetic dataset that provides ground truth annotations for semantic segmentation, instance segmentation, depth estimation, and optical flow for the entire dataset. It consists of five scenes named \textit{"Scene01"}, \textit{"Scene02"}, \textit{"Scene06"}, \textit{"Scene18"}, and \textit{"Scene20"} which account for a total of $2126$ unique frames of stereo images that are augmented to recreate $10$ different environment conditions: clone, fog, morning, overcast, rain, sunset, and four angle variations corresponding to $\pm 15^{\circ}$  and $\pm 30^{\circ}$  around the vertical axis. All in all, Virtual KITTI 2 contains $21260$ RGB stereo frames. 

In our experiments, we discarded the angle variation splits, $\pm 15^{\circ}$  and $\pm 30^{\circ}$, to reduce redundancy in the dataset and kept the other six splits for training, evaluation, and testing. We trained on scenes \textit{"Scene01"}, \textit{"Scene06"}, and \textit{"Scene20"}, evaluated on \textit{"Scene18"} and tested on \textit{"Scene02"}. We resized the input frames to $200px$ height and $1000px$ width.

\begin{figure}
     \centering
     \begin{subfigure}[b]{0.45\textwidth}
         \centering
         \includegraphics[width=\textwidth]{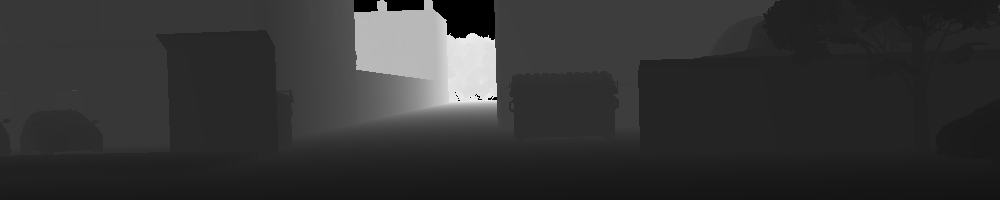}
         \caption{Fully dense depth map.}
         \vspace{2mm}
         \label{fig:depth100}
     \end{subfigure}
     
     \begin{subfigure}[b]{0.45\textwidth}
         \centering
         \includegraphics[width=\textwidth]{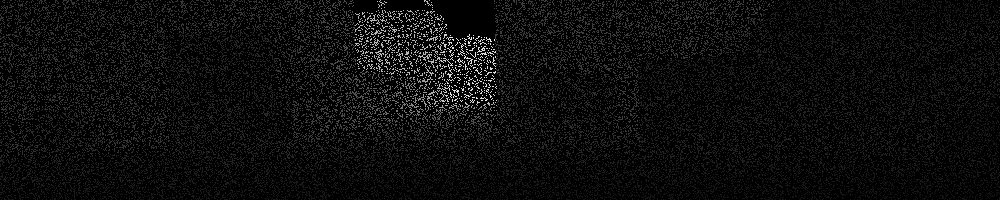}
         \caption{Depth map, $sparsity=20\%$.}
         \vspace{2mm}
         \label{fig:depth25}
     \end{subfigure}
     \begin{subfigure}[b]{0.45\textwidth}
         \centering
         \includegraphics[width=\textwidth]{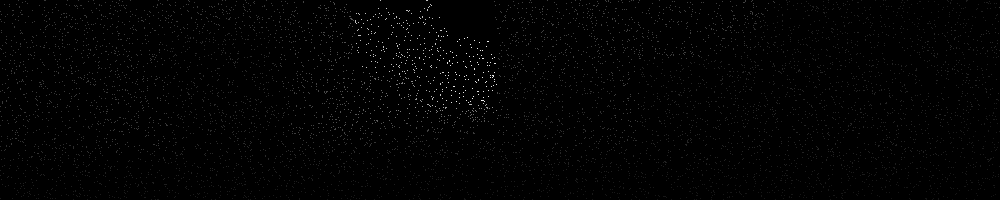}
         \caption{Depth map, $sparsity=5\%$.}
         \vspace{2mm}
         \label{fig:depth5}
     \end{subfigure}
        \caption{Depth maps visualization at different sparsity levels}
        \label{fig:splevel}
\end{figure}

\paragraph{Pre-processing}Since Virtual KITTI 2 is a synthetic dataset, it provides fully dense depth maps for every single frame, however, in order to recreate real conditions as close as possible, we sampled the ground truth maps and set the sparsity to $20\%$, meaning that only $20\%$ of the pixels from any given image would have a depth value available. On the other hand, non-ground-truth maps were sampled to have a sparsity of $5\%$. Under real-world conditions, $3D$ scenes are mapped with laser scanner devices, and when the $3D$ points are projected onto a $2D$ plane, they account for approximately $5\%$ coverage of the entire image. The ground truth, however, is usually obtained by merging consecutive maps together, thus increasing the sparsity to around $20\%$ \citep{sparsity}. Figure \ref{fig:splevel} depicts the visual contrast between different sparsity levels.

\paragraph{Panoptic Segmentation Annotations for Virtual KITTI 2} Although Virtual Kitti 2 does not provide panoptic segmentation annotations directly, it is possible to use semantic segmentation and instance segmentation annotations to generate ground truth panoptic segmentation annotations.  All the scripts are provided in the code repository. Thus, we hope to increase the interest of the community in this dataset as well as other possible datasets for which this approach might be found suitable and useful.

\begin{figure*}[ht]
\centering
\begin{tabular}{cccc}
{{\rotatebox[origin=tl]{90}{(a)}}} &
{\includegraphics[width = 1.8in]{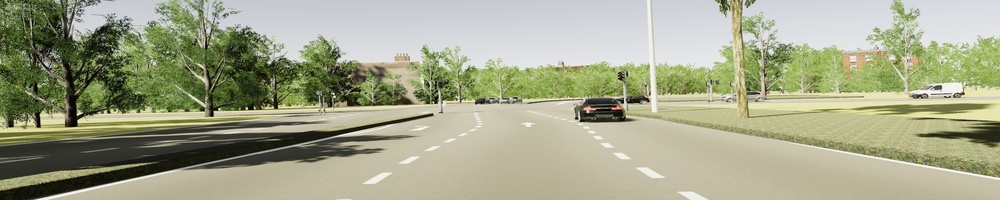}} &
{\includegraphics[width = 1.8in]{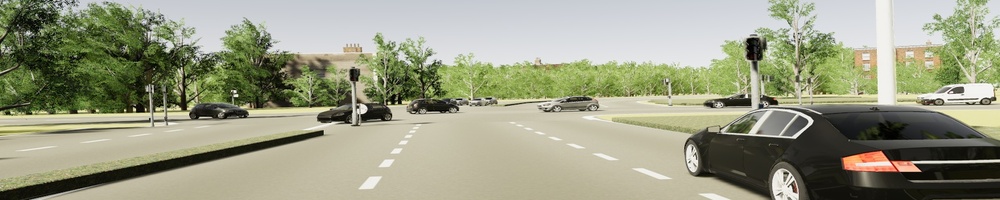}} &
{\includegraphics[width = 1.8in]{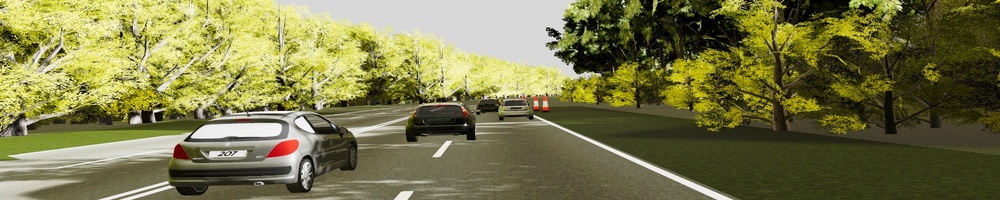}} \\

{{\rotatebox[origin=tl]{90}{(b)}}}&
{\includegraphics[width = 1.8in]{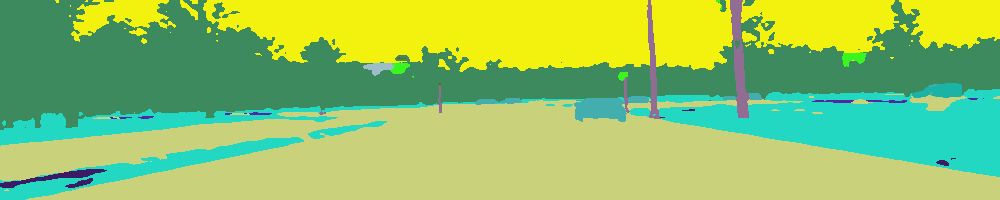}} &
{\includegraphics[width = 1.8in]{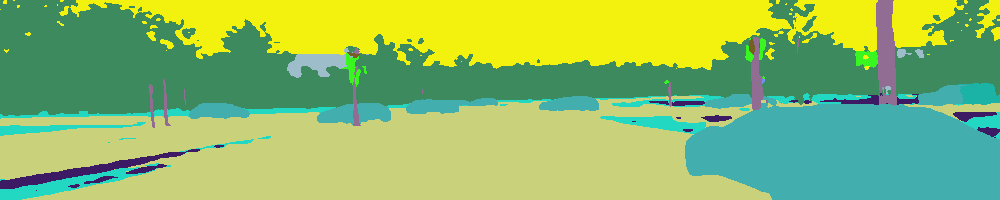}} &
{\includegraphics[width = 1.8in]{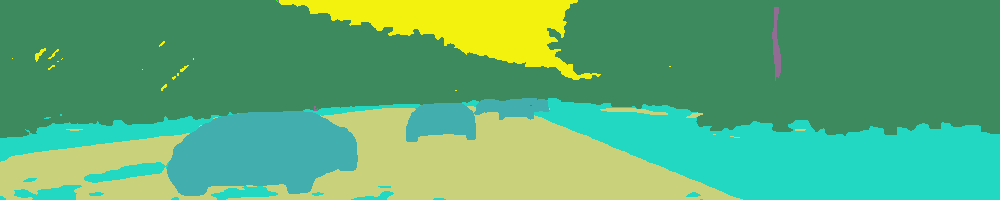}}\\

{{\rotatebox[origin=tl]{90}{(c)}}}&
{\includegraphics[width = 1.8in]{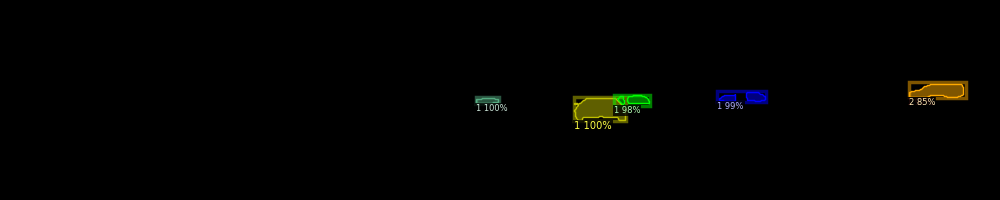}} &
{\includegraphics[width = 1.8in]{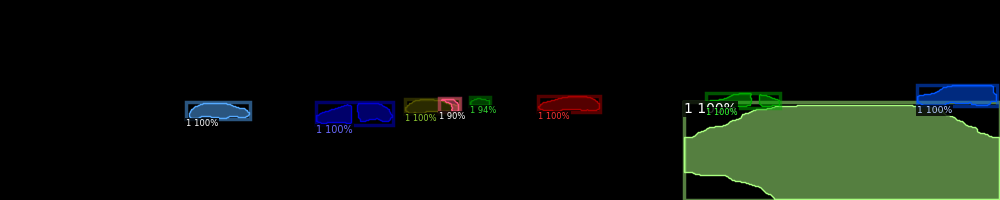}} &
{\includegraphics[width = 1.8in]{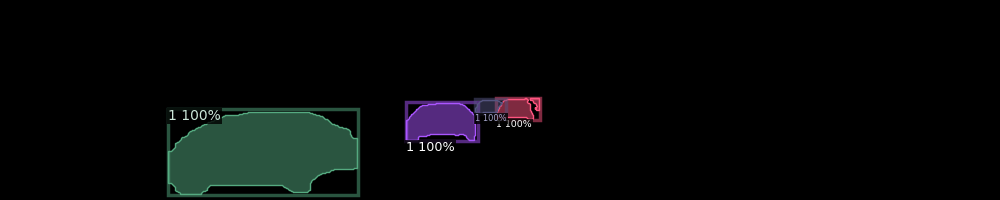}}\\

{{\rotatebox[origin=tl]{90}{(d)}}}&
{\includegraphics[width = 1.8in]{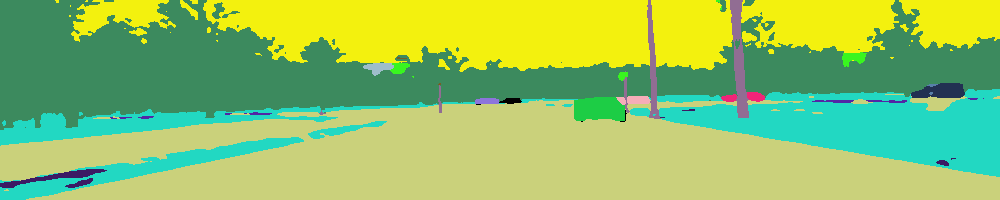}} &
{\includegraphics[width = 1.8in]{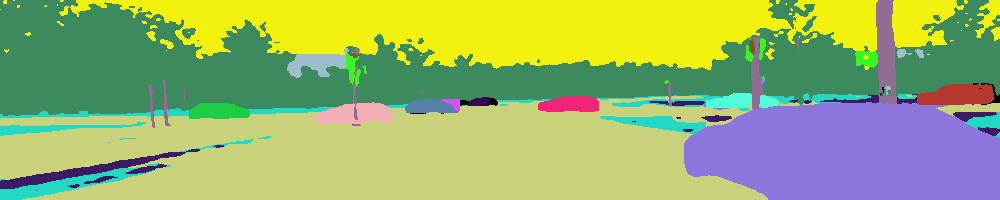}} &
{\includegraphics[width = 1.8in]{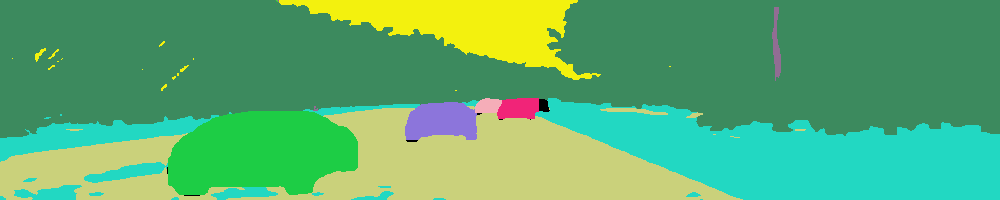}}\\

{{\rotatebox[origin=tl]{90}{(e)}}}&
{\includegraphics[width = 1.8in]{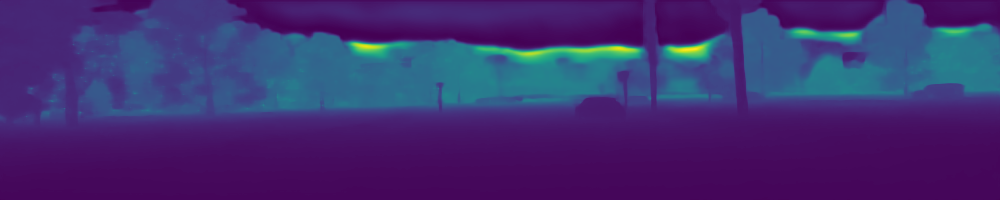}} &
{\includegraphics[width = 1.8in]{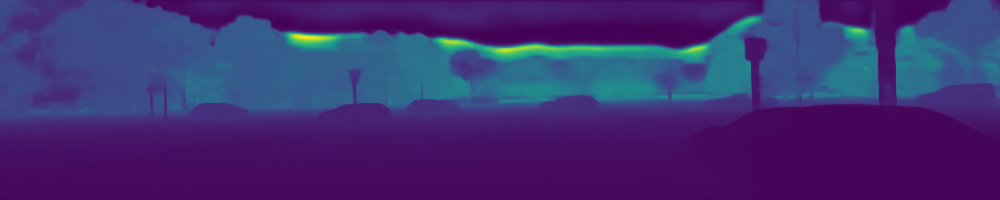}} &
{\includegraphics[width = 1.8in]{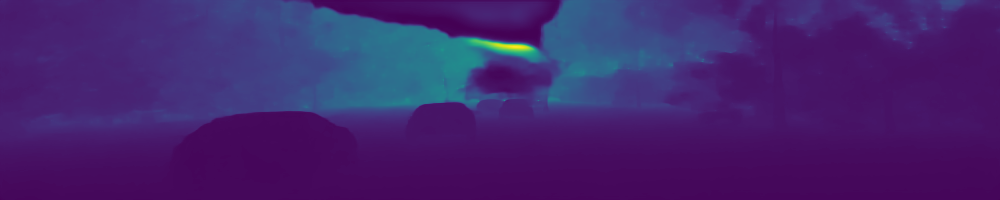}}\\

{{\rotatebox[origin=tl]{90}{(f)}}}&
{\includegraphics[width = 1.8in]{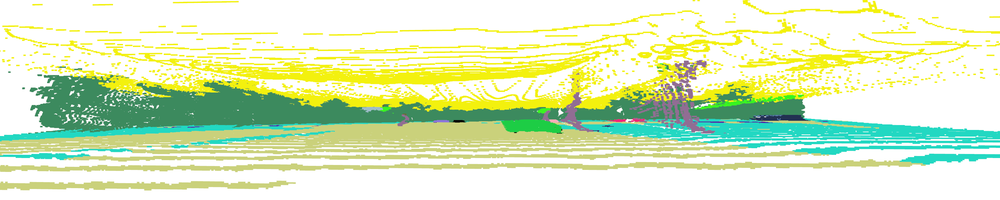}} &
{\includegraphics[width = 1.8in]{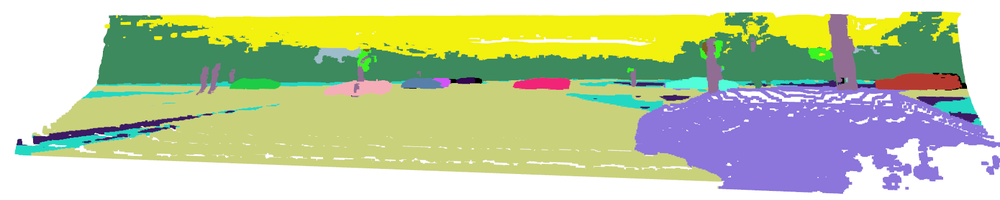}} &
{\includegraphics[width = 1.8in]{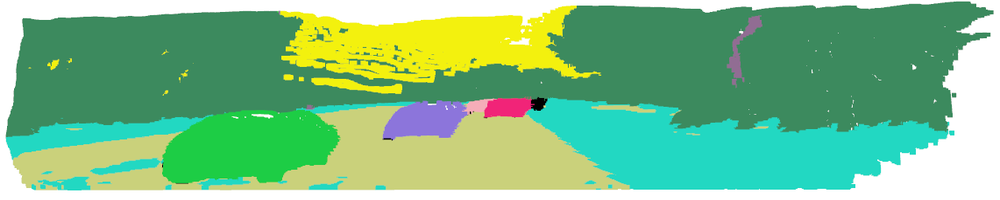}}\\

\end{tabular}
\caption{ Panoptic segmentation and depth completion results on Virtual KITTI 2. Rows from top down show:   \textit{(a)} RGB input images, \textit{(b)} semantic segmentation, \textit{(c)} instance segmentation, \textit{(d)} panoptic segmentation, \textit{(e)} depth completion output, and \textit{(f)} 3D panoptic segmentation.}
\label{pandepth_res}
\end{figure*}

\begin{table*}
\caption{Results of our model compared to baselines.} \centering
\begin{tabular}{|c|c|c|c|c|c|c|c|c|}
  \hline
  Method & mIoU & mAP  & RMSE(mm) & PQ & RQ & SQ\\
  \hline
  Semantic\_only  & 0.380 & -  & - & -  & -  & -\\
  \hline
  Instance\_only  & - & 0.691 & - & -  & -  & -\\
  \hline
  Depth\_only  & - & -  & 623 & -  & -  & -\\
  \hline
  SemSegDepth & 0.387 & - & 677 & - & - & -\\
  \hline
  \textbf{PanDepth(ours)}  & 0.413 & 0.597  & 653 & 0.384 & 0.450 & 0.467\\
  \hline
\end{tabular}
\label{results1}
\end{table*}

\subsection{Evaluation Metrics}

We calculated the standard COCO metrics \citep{coco} for every task. More specifically, we computed the Intersection over Union (IoU) for semantic segmentation, Mean Average Precision (mAP) for object detection, as well as PQ, recognition quality (RQ), and segmentation quality (SQ) for panoptic segmentation. In addition to the COCO metrics, we computed the root means squared error (RMSE) to evaluate the performance of the depth completion task.

\subsection{Results} \label{results}

We compared the proposed PanDepth model against equivalent models where only one of the task-specific branches of PanDepth is enabled. Such models are listed in Table \ref{results1} as \textit{"Semantic\_only"}, \textit{"Instance\_only"}, and \textit{"Depth\_only"}. Table \ref{results1} also shows the performance of the proposed model PanDepth compared to SemSegDepth \citep{lagos}, a joint-learning model for semantic segmentation and depth completion. SemSegDepth is a multi-task learning model that follows an architecture  similar to that of our model PanDepth. On one hand, it consists of task-specific branches with a shared backbone. On the other hand, the input comprises heterogeneous data, namely, RGB frames and sparse depth maps. However, our model solves more tasks, thus providing a more holistic representation of the input scenes, while keeping high accuracy in all evaluation metrics as shown in Table \ref{results1}. The qualitative results of the proposed model can be inspected visually in Figure \ref{pandepth_res}, where the output of every individual task is depicted as well as a 3D panoptic segmentation reconstructed using the corresponding depth completion output and panoptic segmentation output.

Our model outperforms SemSegDepth in both the accuracy of the semantic segmentation task, as measured by the mIoU metric,  and the depth completion task, as measured by the RMSE metric. The proposed PanDepth model also outperforms the semantic-segmentation-only model (\textit{"Semantic\_only"}) providing more evidence of the advantages of joint-learning. Although the single-task models \textit{"Instance\_only"} and \textit{"Depth\_only"}, for instance segmentation and depth completion respectively, show an increase in accuracy compared to PanDepth, as reported by the mAP and the RMSE, the proposed PanDepth model provides a more complete scene understanding of 3D environments which is a favorable trade-off in autonomous driving applications where holistic scene representations are highly valuable.

It is also important to note that the size of our model does not increase significantly despite solving multiple tasks. That is due to sharing structures such as the feature extractor and relatively small model branches as shown in  Table \ref{tab:model_size}.

\begin{table}[ht]
\caption{Model size.}\label{tab:model_size} \centering
\begin{tabular}{|c|c|c|c|}
  \hline
  Structure & Params\\
  \hline
  Backbone (EfficientNet-B5 ) & 25.2M \\
  \hline
  2-way FPN  & 1.5M\\
  \hline
  Semantic Branch  & 1.2M\\
  \hline
  Instance Branch  & 53.1M \\
  \hline
  Depth Branch  & 1.9M\\
  \hline
  Joint Branch  & 1.2M\\
  \hline
  \textbf{PanDepth Total Params}  & 84M\\
  \hline
\end{tabular}
\end{table}

\section{\uppercase{Conclusions}}

This paper presents an end-to-end model for panoptic segmentation and depth completion using heterogeneous data as input, namely RGB images, and sparse depth maps. Our model yields a better scene understanding by providing a semantic representation of 3D environments. We propose a joint-learning  method to perform multiple tasks, specifically semantic segmentation, instance segmentation, depth completion, and panoptic segmentation. Through a rigorous set of experiments, we demonstrate, quantitatively and qualitatively, the advantages of joint learning and multi-task models. Our model solves multiple computer vision tasks, keeping high-accuracy results compared to other strong baselines, without a significant increase in computational cost.

\label{sec:conclusion}

\section{Acknowledgements}
This work is supported by the Academy of Finland (projects 327910 \& 324346).

\bibliographystyle{apalike}
{\small
\bibliography{Example}}

\begin{thebibliography}{}

\bibitem[Adelson, 2001]{Adelson2001OnSS}
Adelson, E.~H. (2001).
\newblock On seeing stuff: the perception of materials by humans and machines.
\newblock In {\em IS\&T/SPIE Electronic Imaging}.

\bibitem[Bul{\`{o}} et~al., 2017]{iabn}
Bul{\`{o}}, S.~R., Porzi, L., and Kontschieder, P. (2017).
\newblock In-place activated batchnorm for memory-optimized training of dnns.
\newblock {\em CoRR}, abs/1712.02616.

\bibitem[Cabon et~al., 2020]{cabon2020virtual}
Cabon, Y., Murray, N., and Humenberger, M. (2020).
\newblock Virtual kitti 2.

\bibitem[Carion et~al., 2020]{detr}
Carion, N., Massa, F., Synnaeve, G., Usunier, N., Kirillov, A., and Zagoruyko,
  S. (2020).
\newblock End-to-end object detection with transformers.
\newblock {\em CoRR}, abs/2005.12872.

\bibitem[Chen et~al., 2020a]{sc}
Chen, L., Wang, H., and Qiao, S. (2020a).
\newblock Scaling wide residual networks for panoptic segmentation.
\newblock {\em CoRR}, abs/2011.11675.

\bibitem[Chen et~al., 2020b]{2D3D}
Chen, Y., Yang, B., Liang, M., and Urtasun, R. (2020b).
\newblock Learning joint 2d-3d representations for depth completion.
\newblock {\em CoRR}, abs/2012.12402.

\bibitem[Chen et~al., 2020c]{chen2020learning}
Chen, Y., Yang, B., Liang, M., and Urtasun, R. (2020c).
\newblock Learning joint 2d-3d representations for depth completion.

\bibitem[Cheng et~al., 2019]{panoptic_deeplab}
Cheng, B., Collins, M.~D., Zhu, Y., Liu, T., Huang, T.~S., Adam, H., and Chen,
  L. (2019).
\newblock Panoptic-deeplab: {A} simple, strong, and fast baseline for bottom-up
  panoptic segmentation.
\newblock {\em CoRR}, abs/1911.10194.

\bibitem[Cheng et~al., 2021a]{mask2former}
Cheng, B., Misra, I., Schwing, A.~G., Kirillov, A., and Girdhar, R. (2021a).
\newblock Masked-attention mask transformer for universal image segmentation.
\newblock {\em CoRR}, abs/2112.01527.

\bibitem[Cheng et~al., 2021b]{maskt}
Cheng, B., Schwing, A.~G., and Kirillov, A. (2021b).
\newblock Per-pixel classification is not all you need for semantic
  segmentation.
\newblock {\em CoRR}, abs/2107.06278.

\bibitem[Chollet, 2016]{xception}
Chollet, F. (2016).
\newblock Xception: Deep learning with depthwise separable convolutions.
\newblock {\em CoRR}, abs/1610.02357.

\bibitem[de~Geus et~al., 2019]{fast}
de~Geus, D., Meletis, P., and Dubbelman, G. (2019).
\newblock Fast panoptic segmentation network.
\newblock {\em CoRR}, abs/1910.03892.

\bibitem[Eldesokey et~al., 2018]{confprop}
Eldesokey, A., Felsberg, M., and Khan, F.~S. (2018).
\newblock Confidence propagation through cnns for guided sparse depth
  regression.
\newblock {\em CoRR}, abs/1811.01791.

\bibitem[Gansbeke et~al., 2019]{spnoisy}
Gansbeke, W.~V., Neven, D., Brabandere, B.~D., and Gool, L.~V. (2019).
\newblock Sparse and noisy lidar completion with {RGB} guidance and
  uncertainty.
\newblock {\em CoRR}, abs/1902.05356.

\bibitem[Gao et~al., 2022]{PanopticDepth}
Gao, N., He, F., Jia, J., Shan, Y., Zhang, H., Zhao, X., and Huang, K. (2022).
\newblock Panopticdepth: A unified framework for depth-aware panoptic
  segmentation.

\bibitem[Guo et~al., 2020]{ltb}
Guo, P., Lee, C., and Ulbricht, D. (2020).
\newblock Learning to branch for multi-task learning.
\newblock {\em CoRR}, abs/2006.01895.

\bibitem[Hazirbas et~al., 2016]{hazirbas16fusenet}
Hazirbas, C., Ma, L., Domokos, C., and Cremers, D. (2016).
\newblock Fusenet: Incorporating depth into semantic segmentation via
  fusion-based cnn architecture.
\newblock In {\em Asian Conference on Computer Vision (ACCV)}.

\bibitem[He et~al., 2017]{maskrcnn}
He, K., Gkioxari, G., Doll{\'{a}}r, P., and Girshick, R.~B. (2017).
\newblock Mask {R-CNN}.
\newblock {\em CoRR}, abs/1703.06870.

\bibitem[He et~al., 2021]{SOSD}
He, L., Lu, J., Wang, G., Song, S., and Zhou, J. (2021).
\newblock Sosd-net: Joint semantic object segmentation and depth estimation
  from monocular images.
\newblock {\em CoRR}, abs/2101.07422.

\bibitem[Hou et~al., 2019]{rt_pan}
Hou, R., Li, J., Bhargava, A., Raventos, A., Guizilini, V., Fang, C., Lynch,
  J.~P., and Gaidon, A. (2019).
\newblock Real-time panoptic segmentation from dense detections.
\newblock {\em CoRR}, abs/1912.01202.

\bibitem[Hu et~al., 2021]{PEnet}
Hu, M., Wang, S., Li, B., Ning, S., Fan, L., and Gong, X. (2021).
\newblock Penet: Towards precise and efficient image guided depth completion.
\newblock {\em CoRR}, abs/2103.00783.

\bibitem[Jaritz et~al., 2018]{depthandsem}
Jaritz, M., de~Charette, R., Wirbel, {\'{E}}., Perrotton, X., and Nashashibi,
  F. (2018).
\newblock Sparse and dense data with cnns: Depth completion and semantic
  segmentation.
\newblock {\em CoRR}, abs/1808.00769.

\bibitem[Kirillov et~al., 2019]{panoptic_fpn}
Kirillov, A., Girshick, R.~B., He, K., and Doll{\'{a}}r, P. (2019).
\newblock Panoptic feature pyramid networks.
\newblock {\em CoRR}, abs/1901.02446.

\bibitem[Kirillov et~al., 2018]{panoptic}
Kirillov, A., He, K., Girshick, R.~B., Rother, C., and Doll{\'{a}}r, P. (2018).
\newblock Panoptic segmentation.
\newblock {\em CoRR}, abs/1801.00868.

\bibitem[Lagos and Rahtu, 2022]{lagos}
Lagos, J.~P. and Rahtu, E. (2022).
\newblock Semsegdepth: {A} combined model for semantic segmentation and depth
  completion.
\newblock In Farinella, G.~M., Radeva, P., and Bouatouch, K., editors, {\em
  Proceedings of the 17th International Joint Conference on Computer Vision,
  Imaging and Computer Graphics Theory and Applications, {VISIGRAPP} 2022,
  Volume 5: VISAPP, Online Streaming, February 6-8, 2022}, pages 155--165.
  {SCITEPRESS}.

\bibitem[Li and Dong, 2021]{aux2}
Li, B. and Dong, A. (2021).
\newblock Multi-task learning with attention : Constructing auxiliary tasks for
  learning to learn.
\newblock In {\em 2021 IEEE 33rd International Conference on Tools with
  Artificial Intelligence (ICTAI)}, pages 145--152.

\bibitem[Li et~al., 2018]{attention_guided}
Li, Y., Chen, X., Zhu, Z., Xie, L., Huang, G., Du, D., and Wang, X. (2018).
\newblock Attention-guided unified network for panoptic segmentation.
\newblock {\em CoRR}, abs/1812.03904.

\bibitem[Li et~al., 2021]{pant2}
Li, Z., Wang, W., Xie, E., Yu, Z., Anandkumar, A., Alvarez, J.~M., Lu, T., and
  Luo, P. (2021).
\newblock Panoptic segformer.
\newblock {\em CoRR}, abs/2109.03814.

\bibitem[Liebel and K{\"{o}}rner, 2019]{multidepth}
Liebel, L. and K{\"{o}}rner, M. (2019).
\newblock Multidepth: Single-image depth estimation via multi-task regression
  and classification.
\newblock {\em CoRR}, abs/1907.11111.

\bibitem[Liebel and Körner, 2018a]{liebel2018auxiliary}
Liebel, L. and Körner, M. (2018a).
\newblock Auxiliary tasks in multi-task learning.

\bibitem[Liebel and Körner, 2018b]{auxiliary}
Liebel, L. and Körner, M. (2018b).
\newblock Auxiliary tasks in multi-task learning.

\bibitem[Lin et~al., 2014]{coco}
Lin, T., Maire, M., Belongie, S.~J., Bourdev, L.~D., Girshick, R.~B., Hays, J.,
  Perona, P., Ramanan, D., Doll{\'{a}}r, P., and Zitnick, C.~L. (2014).
\newblock Microsoft {COCO:} common objects in context.
\newblock {\em CoRR}, abs/1405.0312.

\bibitem[Liu et~al., 2019]{end2}
Liu, H., Peng, C., Yu, C., Wang, J., Liu, X., Yu, G., and Jiang, W. (2019).
\newblock An end-to-end network for panoptic segmentation.
\newblock {\em CoRR}, abs/1903.05027.

\bibitem[Liu et~al., 2018]{multiatt}
Liu, S., Johns, E., and Davison, A.~J. (2018).
\newblock End-to-end multi-task learning with attention.
\newblock {\em CoRR}, abs/1803.10704.

\bibitem[Long et~al., 2014]{DBLP}
Long, J., Shelhamer, E., and Darrell, T. (2014).
\newblock Fully convolutional networks for semantic segmentation.
\newblock {\em CoRR}, abs/1411.4038.

\bibitem[Mohan and Valada, 2020]{effps}
Mohan, R. and Valada, A. (2020).
\newblock Efficientps: Efficient panoptic segmentation.
\newblock {\em CoRR}, abs/2004.02307.

\bibitem[Park et~al., 2020]{nonloc}
Park, J., Joo, K., Hu, Z., Liu, C., and Kweon, I.~S. (2020).
\newblock Non-local spatial propagation network for depth completion.
\newblock {\em CoRR}, abs/2007.10042.

\bibitem[Petrovai and Nedevschi, 2019]{autod}
Petrovai, A. and Nedevschi, S. (2019).
\newblock Multi-task network for panoptic segmentation in automated driving.
\newblock In {\em 2019 IEEE Intelligent Transportation Systems Conference
  (ITSC)}, pages 2394--2401.

\bibitem[Qiu et~al., 2018]{deeplidar}
Qiu, J., Cui, Z., Zhang, Y., Zhang, X., Liu, S., Zeng, B., and Pollefeys, M.
  (2018).
\newblock Deeplidar: Deep surface normal guided depth prediction for outdoor
  scene from sparse lidar data and single color image.
\newblock {\em CoRR}, abs/1812.00488.

\bibitem[Ronneberger et~al., 2015]{unet}
Ronneberger, O., Fischer, P., and Brox, T. (2015).
\newblock U-net: Convolutional networks for biomedical image segmentation.
\newblock {\em CoRR}, abs/1505.04597.

\bibitem[Ruder, 2017]{mtask}
Ruder, S. (2017).
\newblock An overview of multi-task learning in deep neural networks.
\newblock {\em CoRR}, abs/1706.05098.

\bibitem[Schon et~al., 2021]{Schon_2021}
Schon, M., Buchholz, M., and Dietmayer, K. (2021).
\newblock {MGNet}: Monocular geometric scene understanding for autonomous
  driving.
\newblock In {\em 2021 {IEEE}/{CVF} International Conference on Computer Vision
  ({ICCV})}. {IEEE}.

\bibitem[Sener and Koltun, 2018]{mtask2}
Sener, O. and Koltun, V. (2018).
\newblock Multi-task learning as multi-objective optimization.
\newblock {\em CoRR}, abs/1810.04650.

\bibitem[Tan and Le, 2019]{effnet}
Tan, M. and Le, Q.~V. (2019).
\newblock Efficientnet: Rethinking model scaling for convolutional neural
  networks.
\newblock {\em CoRR}, abs/1905.11946.

\bibitem[Tang et~al., 2019]{guidedconv}
Tang, J., Tian, F., Feng, W., Li, J., and Tan, P. (2019).
\newblock Learning guided convolutional network for depth completion.
\newblock {\em CoRR}, abs/1908.01238.

\bibitem[Uhrig et~al., 2017]{sparsity}
Uhrig, J., Schneider, N., Schneider, L., Franke, U., Brox, T., and Geiger, A.
  (2017).
\newblock Sparsity invariant cnns.
\newblock {\em CoRR}, abs/1708.06500.

\bibitem[Vaswani et~al., 2017]{attention}
Vaswani, A., Shazeer, N., Parmar, N., Uszkoreit, J., Jones, L., Gomez, A.~N.,
  Kaiser, L., and Polosukhin, I. (2017).
\newblock Attention is all you need.
\newblock {\em CoRR}, abs/1706.03762.

\bibitem[Wang et~al., 2020]{pant1}
Wang, H., Zhu, Y., Adam, H., Yuille, A.~L., and Chen, L. (2020).
\newblock Max-deeplab: End-to-end panoptic segmentation with mask transformers.
\newblock {\em CoRR}, abs/2012.00759.

\bibitem[Wang et~al., 2018]{Wang_2018}
Wang, S., Suo, S., Ma, W.-C., Pokrovsky, A., and Urtasun, R. (2018).
\newblock Deep parametric continuous convolutional neural networks.
\newblock {\em 2018 IEEE/CVF Conference on Computer Vision and Pattern
  Recognition}.

\bibitem[Xiong et~al., 2019]{UPS}
Xiong, Y., Liao, R., Zhao, H., Hu, R., Bai, M., Yumer, E., and Urtasun, R.
  (2019).
\newblock Upsnet: {A} unified panoptic segmentation network.
\newblock {\em CoRR}, abs/1901.03784.

\bibitem[Yang et~al., 2019]{DDP}
Yang, Y., Wong, A., and Soatto, S. (2019).
\newblock Dense depth posterior {(DDP)} from single image and sparse range.
\newblock {\em CoRR}, abs/1901.10034.

\bibitem[Yuan et~al., 2021]{PolyphonicFormer}
Yuan, H., Li, X., Yang, Y., Cheng, G., Zhang, J., Tong, Y., Zhang, L., and Tao,
  D. (2021).
\newblock Polyphonicformer: Unified query learning for depth-aware video
  panoptic segmentation.
\newblock {\em CoRR}, abs/2112.02582.

\bibitem[Zhu et~al., 2020]{detr2}
Zhu, X., Su, W., Lu, L., Li, B., Wang, X., and Dai, J. (2020).
\newblock Deformable {DETR:} deformable transformers for end-to-end object
  detection.
\newblock {\em CoRR}, abs/2010.04159.

\bibitem[Zou et~al., 2020a]{simsemseg}
Zou, N., Xiang, Z., Chen, Y., Chen, S., and Qiao, C. (2020a).
\newblock Simultaneous semantic segmentation and depth completion with
  constraint of boundary.
\newblock {\em Sensors}, 20(3).

\bibitem[Zou et~al., 2020b]{Simultaneous}
Zou, N., Xiang, Z., Chen, Y., Chen, S., and Qiao, C. (2020b).
\newblock Simultaneous semantic segmentation and depth completion with
  constraint of boundary.
\newblock {\em Sensors}, 20(3).

\end{thebibliography}

\end{document}